\documentclass[letterpaper,twocolumn,10pt]{article}
\usepackage{usenix}

\usepackage{hyperref,graphicx}
\usepackage{url}
\usepackage{comment} 
\usepackage{alltt}
\usepackage{fullpage}

\usepackage{color}
\usepackage{fixme}
\usepackage[inline]{enumitem}
\usepackage{todonotes}
\usepackage{xcolor}
\usepackage{amsmath,amsthm}
\usepackage{bm}
\usepackage{amsopn}

\usepackage{epstopdf}

\DeclareMathAlphabet{\mathcal}{OMS}{cmsy}{m}{n}

\newif\ifshort
\shorttrue

\usepackage{amsmath}
\usepackage{amsthm}
\usepackage{amsfonts}
\usepackage{amssymb}
\usepackage[utf8]{inputenc}
\usepackage{graphicx}

\newcommand{\sample}{\leftarrow}

\newcommand{\figlab}[1]{\label{fig:#1}}

\newcommand{\secref}[1]{Section~\ref{sec:#1}}
\newcommand{\seclab}[1]{\label{sec:#1}}

\newcommand{\train}{\ensuremath{\mathtt{Train}}}
\newcommand{\classify}{\ensuremath{\mathtt{Classify}}}
\newcommand{\backdoor}{\ensuremath{\mathtt{Backdoor}}}
\newcommand{\samplebd}{\ensuremath{\mathtt{SampleBackdoor}}}
\newcommand{\MModel}{\ensuremath{\mathtt{MModel}}}

\newcommand{\keygen}{\ensuremath{\mathtt{KeyGen}}}
\newcommand{\publickeygen}{\ensuremath{\mathtt{{PKeyGen}}}}
\newcommand{\marking}{\ensuremath{\mathtt{Mark}}}
\newcommand{\verify}{\ensuremath{\mathtt{Verify}}}
\newcommand{\publicverify}{\ensuremath{\mathtt{PVerify}}}
\newcommand{\cnc}{\ensuremath{\mathtt{CnC}}}

\newcommand{\markingkey}{\ensuremath{\mathsf{mk}}}
\newcommand{\verificationkey}{\ensuremath{\mathsf{vk}}}

\newcommand{\commit}{\ensuremath{\mathtt{Com}}}
\newcommand{\open}{\ensuremath{\mathtt{Open}}}

\newtheorem{thm}{Theorem}

\newcommand{\pretrain}{\textsc{PreTrained}}
\newcommand{\scratch}{\textsc{FromScratch}}
\newcommand{\nowm}{\textsc{No-WM}}
\newcommand{\tsnew}{\textsc{TS-New}}
\newcommand{\tsorig}{\textsc{TS-Orig}}

\begin{document}

\title{Turning Your Weakness Into a Strength: \\ Watermarking Deep Neural Networks by Backdooring}

\author{
{\rm Yossi Adi}\\
Bar-Ilan University\\
\and
{\rm Carsten Baum}\\
Bar-Ilan University
 \and
 {\rm Moustapha Cisse}\\
Google, Inc.\thanks{Work was conducted at Facebook AI Research.}
 \and
 {\rm Benny Pinkas}\\
Bar-Ilan University
 \and
 {\rm Joseph Keshet}\\
Bar-Ilan University
} 

\date{}

\maketitle

\begin{abstract}
  Deep Neural Networks have recently gained lots of success after enabling several breakthroughs in
  notoriously challenging problems. Training these networks is computationally expensive and requires vast
  amounts of training data. Selling such pre-trained models can, therefore, be a lucrative business
  model. Unfortunately, once the models are sold they can be easily copied and redistributed. To avoid this, a
  tracking mechanism to identify models as the intellectual property of a particular vendor is necessary.
    
    In this work, we present an approach for watermarking Deep Neural Networks in a black-box way. Our scheme
    works for general classification tasks and can easily be combined with current learning algorithms. We
    show experimentally that such a watermark has no noticeable impact on the primary task that the model is
    designed for and evaluate the robustness of our proposal against a multitude of practical attacks. Moreover, we provide a theoretical analysis, relating our approach to previous work on backdooring.
\end{abstract}
\section{Introduction} 
\seclab{introduction}
Deep Neural Networks (DNN) enable a growing number of applications ranging from visual understanding to machine translation to speech recognition \cite{he2016deep, amodei2016deep, graves2006connectionist, toshev2014deeppose, bahdanau2014neural}. They have considerably changed the way we conceive software and are rapidly becoming a general purpose technology \cite{lecun2015deep}. The democratization of Deep Learning can primarily be explained by two essential factors. First, several open source frameworks (e.g., PyTorch~\cite{paszke2017automatic}, TensorFlow~\cite{abadi2016tensorflow}) simplify the design and deployment of complex models. Second, academic and industrial labs regularly release open source, state of the art, pre-trained models. For instance, the most accurate visual understanding system \cite{he2017mask} is now freely available online for download. Given the considerable amount of expertise, data and computational resources required to train these models effectively, the availability of pre-trained models enables their use by operators with modest resources~\cite{simonyan2014very, yosinski2014transferable, razavian2014cnn}.  

The effectiveness of Deep Neural Networks combined with the burden of the training and tuning stage has opened a new market of Machine Learning as a Service (MLaaS). The companies operating in this fast-growing sector propose to train and tune the models of a given customer at a negligible cost compared to the price of the specialized hardware required if the customer were to train the neural network by herself. Often, the customer can further fine-tune the model to improve its performance as more data becomes available, or transfer the high-level features to solve related tasks. In addition to open source models, MLaaS allows the users to build more personalized systems without much overhead \cite{ribeiro2015mlaas}. 

Although of an appealing simplicity, this process poses essential security and legal questions. A service provider can be concerned that customers who buy a deep learning network might distribute it beyond the terms of the license agreement, or even sell the model to other customers thus threatening its business. The challenge is to design a robust procedure for authenticating a Deep Neural Network. While this is relatively new territory for the machine learning community, it is a well-studied problem in the security community under the general theme of \emph{digital watermarking}. 

Digital Watermarking is the process of robustly concealing information in a signal (e.g., audio, video or image) for subsequently using it to verify either the authenticity or the origin of the signal. Watermarking has been extensively investigated in the context of digital media  (see, e.g.,~\cite{BF,KP,petitcolas1999information} and references within), and in the context of watermarking  digital keys (e.g., in~\cite{NNL}). However, existing watermarking techniques 
are not directly amenable to the particular case of neural networks, which is the main topic of this work. Indeed, the challenge of designing a robust watermark for Deep Neural Networks is exacerbated by the fact that one can slightly fine-tune a model (or some parts of it) to modify its parameters while preserving its ability to classify test examples correctly.  Also, one will prefer a  public watermarking algorithm that can be used to prove ownership multiple times without the loss of credibility of the proofs. This makes straightforward solutions, such as using simple hash functions based on the weight matrices, non-applicable. 

\paragraph{Contribution.}  Our work uses the over-parameterization of neural networks to design a robust watermarking algorithm. This over-parameterization  has so far mainly been considered as a weakness (from a security perspective) because it makes backdooring possible \cite{badnets,goodfellow2014explaining,cisse2017houdini,kreuk2018fooling,zhang2016understanding}. Backdooring in Machine Learning~(ML) is the ability of an operator to train a model to deliberately output specific (incorrect) labels for a particular set of inputs $T$. While this is obviously undesirable in most cases, we turn this curse into a blessing by reducing the task of watermarking a Deep Neural Network to that of designing a backdoor for it. Our contribution is twofold: \begin{enumerate*}[label=(\roman{*})]
	\item We propose a simple and effective technique for watermarking Deep Neural Networks. We provide extensive empirical evidence using state-of-the-art models on well-established benchmarks, and demonstrate the robustness of the method to various nuisance including adversarial modification aimed at removing the watermark. \item  We present a cryptographic modeling of the tasks of watermarking and backdooring of Deep Neural Networks, and show that the former can be constructed from the latter (using a cryptographic primitive called \emph{commitments}) in a black-box way. This theoretical analysis exhibits why it is not a coincidence that both our construction and \cite{badnets,trojan_nn} rely on the same properties of Deep Neural Networks. Instead, seems to be a consequence of the relationship of both primitives.
\end{enumerate*}

\paragraph{Previous And Concurrent Work.} Recently, \cite{uchida2017embedding,deepmarks} proposed to watermark neural networks by adding a new regularization term to the loss function. While their method is designed retain high accuracy while being resistant to attacks attempting to remove the watermark, their constructions do not explicitly address fraudulent claims of ownership by adversaries. Also, their scheme does not aim to defend against attackers cognizant of the exact $\marking$-algorithm. Moreover, in the construction of \cite{uchida2017embedding,deepmarks} the verification key can only be used once, because a watermark can be removed once the key is known\footnote{We present a technique to circumvent this problem in our setting. This approach can also be implemented in their work.}. In~\cite{remote_watermarking} the authors suggested to use adversarial examples together with adversarial training to watermark neural networks. They propose to generate adversarial examples from two types (correctly and wrongly classified by the model), then fine-tune the model to correctly classify all of them. Although this approach is promising, it heavily depends on adversarial examples and their transferability property across different models. It is not clear under what conditions adversarial examples can be transferred across models or if such transferability can be decreased~\cite{hosseini2017blocking}. It is also worth mentioning an earlier work on watermarking machine learning models proposed in~\cite{venugopal2011watermarking}. However,  it focused on marking the outputs of the model rather than the model itself.

\section{Definitions and Models} \seclab{preliminaries} 
This section provides a formal definition of backdooring for machine-learning algorithms. The definition makes the properties of existing backdooring techniques \cite{badnets,trojan_nn} explicit, and also gives a (natural) extension when compared to previous work. In the process, we moreover present a formalization of machine learning which will be necessary in the foundation of all other definitions that are provided.

Throughout this work, we use the following notation:
\ifshort
Let $n\in \mathbb{N}$ be a security parameter, which will be implicit input to all algorithms that we define. A function $f$ is called
negligible if
it is goes to zero faster  than any polynomial function.
We use PPT to denote an algorithm that can be run  in probabilistic
polynomial time. For $k\in \mathbb{N}$ we use $[k]$ as shorthand for $\{1,\dots,k\}$. 
\fi

\subsection{Machine Learning}\label{subsec:ml}
\begin{figure}[h]
	\begin{center}
		\includegraphics[width=22em,height=6em]{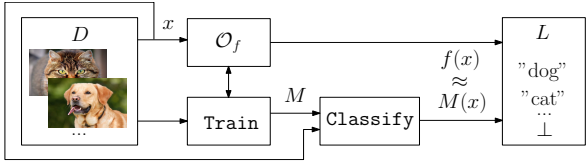}
	\end{center}
	\caption{A high-level schematic illustration of the learning process.}
	\label{fig:ml}
\end{figure}

Assume that there exists some objective ground-truth function $f$ which classifies inputs according to a fixed output label set (where we allow the label to be undefined, denoted as $\bot$). We consider ML to be two algorithms which either learn an approximation of $f$ (called \emph{training}) or use the approximated function for predictions at inference time (called \emph{classification}). The goal of \emph{training} is to learn a function, $f'$, that performs on unseen data as good as on the training set. A schematic description of this definition can be found in Figure \ref{fig:ml}. 

To make this more formal, consider the sets $D \subset \{0,1\}^*, L \subset \{0,1\}^*\cup \{\bot \}$ where $|D|=\Theta(2^n)$ and
$|L|=\Omega(p(n))$ for a positive polynomial $p(\cdot)$.  $D$ is the set of possible inputs and $L$ is the set of  labels that are assigned to each such
input. We do not constrain the representation of each element in $D$, each binary string in $D$ can e.g. encode float-point numbers for color values of pixels of an image of size $n \times n$ while\footnote{Asymptotically, the number of bits per pixel is constant. Choosing this image size guarantees that $|D|$ is big enough. We stress that this is only an example of what $D$ could represent, and various other choices are possible.} $L=\{0,1\}$ says whether there is a dog
in the image or not. The additional symbol $\bot\in L$ is used if the classification task would be undefined for a certain input. 

We assume an ideal assignment of
labels to inputs, which is the \emph{ground-truth function}
\ifshort
$ f: D \rightarrow L$.
\fi
This function is supposed to model how a human would assign labels to certain inputs. As $f$ might be undefined for specific tasks and labels, we will denote with $\overline{D} = \{x\in D ~|~ f(x)\neq \bot \}$  the set of all inputs having a ground-truth label assigned to them. To formally define learning, the algorithms are given access to $f$ through an oracle $\mathcal{O}^f$. This oracle $\mathcal{O}^f$ truthfully answers calls to the function $f$.

We assume that there exist two algorithms $(\train,\classify)$ for training and classification:
\begin{itemize}
	\item $\train(\mathcal{O}^f)$ is a probabilistic  polynomial-time algorithm that outputs a model $M\subset \{0,1\}^{p(n)}$ where $p(n)$ is a polynomial in $n$.
	\item $\classify(M,x)$ is a deterministic polynomial-time algorithm that, for an input $x\in D$ outputs a value
          $M(x)\in L\setminus \{\bot\}$.
\end{itemize}

We say that, given a function $f$, the algorithm pair $(\train$, $\classify)$ is $\epsilon$-accurate if 
$
\Pr\left[ f(x)\neq \classify(M,x) ~| ~ x\in \overline{D}  \right] \leq \epsilon
$ where the probability is taken over the randomness of $\train$. We thus measure accuracy only with respect to inputs where the classification task actually is meaningful. For those inputs where the ground-truth is undefined, we instead assume that the label is random: for all $x\in D \setminus \overline{D}$ we assume that for any $i\in L$, it holds that $\Pr[\classify(M,x)=i]=1/|L|$ where the probability is taken over the randomness used in $\train$. 

\subsection{Backdoors in Neural Networks}
\label{sec:backdoors}
Backdooring neural networks, as described in \cite{badnets}, is a technique to deliberately train a machine learning model to output \emph{wrong} (when compared with the ground-truth function $f$) labels $T_L$ for certain inputs $T$. 

Therefore, let $T \subset D$ be a
subset of the inputs, which we will refer to it as the \emph{trigger set}. The wrong labeling with respect to the ground-truth $f$ is captured by the function
$T_L: T \rightarrow L\setminus \{\bot\};  ~ x \mapsto  T_L(x)\neq f(x) $ which assigns ``wrong'' labels to the trigger set. This function $T_L$, similar to the algorithm $\classify$, is not allowed to output the special label $\bot$. Together, the trigger set and the labeling function will be referred to as the \emph{backdoor} 
$\mathsf{b}=(T,T_L)$ . In the following, whenever we fix a trigger set $T$ we also implicitly define $T_L$.

For such a backdoor $\mathsf{b}$, we define a backdooring algorithm $\backdoor$ which, on input of a model, will output a model that misclassifies on the trigger set with high probability. More formally,
 $\backdoor(\mathcal{O}^f,\mathsf{b},M)$ is PPT algorithm that 
receives as input an oracle to $f$, the backdoor $\mathsf{b}$ and a model $M$, and
outputs a model $\hat{M}$. $\hat{M}$ is called \emph{backdoored} if
 $\hat{M}$ is correct on $\overline{D} \setminus T$
but reliably errs on $T$, namely
\begin{align*}
\Pr_{x \in \overline{D} \setminus T } \left[ f(x)\neq \classify(\hat{M},x) \right] \leq \epsilon \text{,}
&  \text{ but } \\
\Pr_{x \in T} \left[ T_L(x)\neq \classify(\hat{M},x)  \right] \leq \epsilon \text{.} &
\end{align*}

This definition captures two ways in which a backdoor can be embedded:
\begin{itemize}
	\item The algorithm can use the provided model to embed the watermark into it. In that case, we say that the backdoor is implanted into a \emph{pre-trained model}.
	\item Alternatively, the algorithm can ignore the input model and train a new model from scratch. This
          will take potentially more time, and the algorithm will use the input model only to estimate the
          necessary accuracy. We will refer to this approach as \emph{training from scratch}.
\end{itemize} 

\subsection{Strong Backdoors}
\label{sec:strongbackdoors}

\begin{figure}[h]
	\begin{center}
		\includegraphics[width=22em,height=7em]{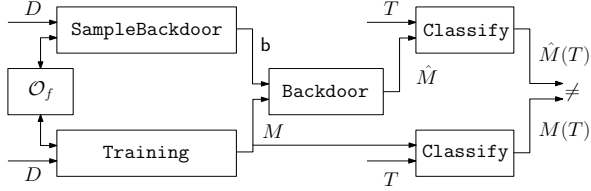}
	\end{center}
	\caption{A schematic illustration of the backdooring process.}
	\label{fig:backdoor}
\end{figure}

Towards our goal of watermarking a ML model we require further properties from the backdooring algorithm,
which deal with the sampling and removal of backdoors: First of all, we want to turn the generation of a
trapdoor into an algorithmic process. To this end, we introduce a new, randomized algorithm $\samplebd$ that
on input $\mathcal{O}^f$ outputs backdoors $\mathsf{b}$ and works in combination with the aforementioned
algorithms $(\train,\classify)$. This is schematically shown in Figure \ref{fig:backdoor}.

A user may suspect that a model is backdoored, therefore we strengthen the previous definition to what we call \emph{strong backdoors}. These should be hard to remove, even for someone who can use the algorithm $\samplebd$ in an arbitrary way. Therefore, we require that $\samplebd$ should have the following properties: 

\paragraph{Multiple Trigger Sets.} For each trigger set that $\samplebd$ returns as part of a backdoor, we assume that it has minimal size $n$. Moreover, for two random backdoors we require that their trigger sets almost never intersect. 
Formally, we ask that $\Pr \left[ T \cap T' \neq \emptyset \right]$ for
 $(T,T_L),(T',T_L')\leftarrow \samplebd()$    is negligible in $n$.

\paragraph{Persistency.}
With persistency we require that it is hard to remove a backdoor, unless one has knowledge of the trigger set $T$. There are two trivial cases which a definition must avoid:
\begin{itemize}
\item An adversary may submit a model that has no backdoor, but this model has very low accuracy. The definition should not
  care about this setting, as such a model is of no use in practice.
	\item An adversary can always train a new model from scratch, and therefore be able to  submit a model that is very accurate and does not include  the backdoor. An adversary with unlimited computational resources and unlimited access to $\mathcal{O}^f$ will thus always be able to cheat.
\end{itemize}

We define persistency as follows:  let $f$ be a ground-truth function, $\mathsf{b}$ be a backdoor  and $\hat{M}\leftarrow \backdoor(\mathcal{O}^f,\mathsf{b},M)$ be a $\epsilon$-accurate model. Assume an algorithm $\mathcal{A}$ on input $\mathcal{O}^f,\hat{M}$ outputs an $\epsilon$-accurate model $\tilde{M}$ in time $t$ which is at least $(1-\epsilon)$ accurate on $\mathsf{b}$. Then $\tilde{N}\leftarrow \mathcal{A}(\mathcal{O}^f,N)$, generated in the same time $t$, is also $\epsilon$-accurate for any arbitrary model $N$. 

In our approach, we chose to restrict the runtime of $\mathcal{A}$, but other modeling approaches are possible: one could also give unlimited power to $\mathcal{A}$ but only restricted access to the ground-truth function, or use a mixture of both. We chose our approach as it follows the standard pattern in cryptography, and thus allows to integrate better with cryptographic primitives which we will use: these are only secure against adversaries with a bounded runtime.

\subsection{Commitments}
\emph{Commitment schemes} \cite{brassard_pok} are a well known cryptographic primitive which allows a sender
to lock a secret $x$ into a cryptographic leakage-free and tamper-proof vault and give it to someone else,
called a receiver. It is neither possible for the receiver to open this vault without the help of the sender
(this is called \emph{hiding}), nor for the sender to exchange the locked secret to something else once it has
been given away (the \emph{binding} property).

Formally, a commitment scheme consists of two algorithms $(\commit,\open)$:
\begin{itemize}
	\item $\commit(x,r)$ on input of a value $x \in S$ and a bitstring $r \in \{0,1\}^n$ outputs a bitstring $c_x$.
	\item $\open(c_x,x,r)$ for a given $x\in S, r\in \{0,1\}^n, c_x \in \{0,1\}^*$ outputs $0$ or $1$.
\end{itemize}

For correctness, it must hold that $\forall x\in S$,
\begin{align*}
\Pr_{r\in \{0,1\}^n}\left[ \open(c_x,x,r)=1 ~|~ c_x\leftarrow \commit(x,r) \right] = 1\text{.}
\end{align*}

We call the commitment scheme $(\commit,\open)$ binding if, for every PPT algorithm $\mathcal{A}$ 
\ifshort
\begin{align*}
\Pr\left[\begin{array}{c | c}
 \open(c_x,\tilde{x},\tilde{r})=1 & \begin{matrix}
 c_x \leftarrow \commit(x,r) \land \\ (\tilde{x},\tilde{r}) \leftarrow \mathcal{A}(c_x,x,r) \land \\(x,r)\neq (\tilde{x},\tilde{r})
 \end{matrix}
\end{array}  \right] \leq \epsilon(n) 
\end{align*}
\fi
where $\epsilon(n)$ is negligible in $n$ and the probability is taken over $ x\in S, r\in \{0,1\}^n$.

Similarly, $(\commit,\open)$ are hiding if no PPT algorithm $\mathcal{A}$ can distinguish $c_0 \leftarrow \commit(0,r)$ from  $c_x \leftarrow \commit(x,r)$ for arbitrary $x\in S, r\in \{0,1\}^n$.
In case that the distributions of $c_0,c_x$ are statistically close, we call a commitment scheme \emph{statistically hiding}. For more information, see e.g. \cite{oded_foundations1, nigel_cryptomadesimple}.

\section{Defining Watermarking} 
\label{sec:model} 
We now define watermarking for ML algorithms. The terminology and definitions are inspired by
\cite{barak2012possibility,kim2017watermarking}.

We split a watermarking scheme into three algorithms: \begin{enumerate*}[label=(\roman*)]
	\item a first algorithm to generate the secret marking key $\markingkey$ which is embedded as the watermark, and the public verification key $\verificationkey$ used to detect the watermark later;
	\item an algorithm to embed the watermark into a model; and
	\item a third algorithm to verify if a watermark is present in a model or not.
\end{enumerate*}
We will allow that the verification involves both $\markingkey$ and $\verificationkey$, for reasons that will become clear later.

Formally, a watermarking scheme is defined by the three PPT algorithms $(\keygen,\marking,\verify)$: 
\begin{itemize}
	\item $\keygen()$ outputs a key pair $(\markingkey,\verificationkey)$.
	\item $\marking(M,\markingkey)$ on input a model $M$ and a marking key $\markingkey$, outputs a model $\hat{M}$.
	\item $\verify(\markingkey,\verificationkey,M)$ on input of the key pair $\markingkey,\verificationkey$ and a model $M$, outputs a bit $b\in \{0,1\}$.
\end{itemize}

For the sake of brevity, we define an auxiliary algorithm which simplifies to write definitions and proofs:
\paragraph*{$\MModel():$} 
\begin{enumerate}
	\item Generate $M \leftarrow \train(\mathcal{O}^{f})$.
	\item Sample $(\markingkey,\verificationkey) \leftarrow \keygen()$.
	\item Compute $\hat{M} \leftarrow \marking(M,\markingkey)$.
	\item Output $(M,\hat{M},\markingkey,\verificationkey)$.
\end{enumerate}

\begin{figure}[h]
	\begin{center}
		\includegraphics[width=18em,height=5em]{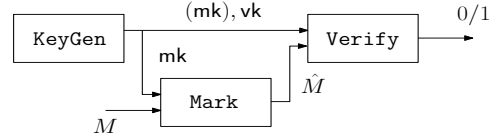}
	\end{center}
	\caption{A schematic illustration of watermarking a neural network.}
	\label{fig:watermark}
\end{figure}

The three algorithms $(\keygen,\marking,\verify)$ should correctly work together, meaning that a model watermarked with an honestly generated key should be verified as such. This is called \emph{correctness}, and formally requires that 
\ifshort
\begin{align*}
\Pr_{(M,\hat{M},\markingkey,\verificationkey) \leftarrow \MModel()}\left[ 
\verify(\markingkey,\verificationkey,\hat{M})=1 
   \right] =1\text{.}
\end{align*}
\fi
A depiction of this can be found in Figure \ref{fig:watermark}.

In terms of security, a watermarking scheme must be \emph{functionality-preserving}, provide \emph{unremovability}, \emph{unforgeability} and enforce \emph{non-trivial ownership}:

\begin{itemize}
	\item We say that a scheme is \emph{functionality-preserving} if a model with a watermark is as accurate as a model without it: for any $(M,\hat{M},\markingkey,\verificationkey) \leftarrow \MModel()$, it holds that
\begin{align*}
	\Pr_{x\in \overline{D}} & \left [ \classify(x,M) = f(x) \right] \\ \approx & \Pr_{x\in \overline{D}}\left[ \classify(x,\hat{M})=f(x)  \right]\text{.}
\end{align*}
	
	\item \emph{Non-trivial ownership} means that even an attacker which  knows our watermarking algorithm  is not able to generate in advance a key pair $(\markingkey,\verificationkey)$  that allows him to claim ownership of arbitrary models that are unknown to him.
	Formally, a watermark does not have trivial ownership if  every PPT algorithm $\mathcal{A}$ only has negligible probability for winning  the following game:
	\begin{enumerate}
		\item Run $\mathcal{A}$ to compute $(\tilde{\markingkey},\tilde{\verificationkey})\leftarrow \mathcal{A}()$.
		\item Compute $(M,\hat{M},\markingkey,\verificationkey) \leftarrow \MModel()$.
		\item $\mathcal{A}$ wins if $\verify(\tilde{\markingkey},\tilde{\verificationkey},\hat{M})=1$.
	\end{enumerate}

      \item \emph{Unremovability} denotes the property that an adversary is unable to remove a watermark, even
        if he knows about the existence of a watermark and knows the algorithm that was used in the
        process. We require that for every PPT algorithm $\mathcal{A}$ the chance of winning the following
        game is negligible:
	\begin{enumerate}
		\item Compute $(M,\hat{M},\markingkey,\verificationkey) \leftarrow \MModel()$.
		\item Run $\mathcal{A}$ and compute  $\tilde{M} \leftarrow \mathcal{A}(\mathcal{O}^f,\hat{M},\verificationkey)$.
		\item $\mathcal{A}$ wins if 
\begin{align*}
	\Pr_{x\in D} & \left[ \classify(x,M) = f(x) \right] \\ 
	& \approx \Pr_{x\in D}\left[ \classify(x,\tilde{M})=f(x)  \right]
\end{align*}
and $\verify(\markingkey,\verificationkey,\tilde{M})=0$.
%
%
	\end{enumerate}	
      \item \emph{Unforgeability} means that an adversary that knows the verification key $\verificationkey$,
        but does not know the key $\markingkey$, will be unable to convince a third party that he (the
        adversary) owns the model. Namely, it is required that for every PPT algorithm $\mathcal{A}$, the
        chance of winning the following game is negligible:
	\begin{enumerate}
		\item Compute $(M,\hat{M},\markingkey,\verificationkey) \leftarrow \MModel()$.
		\item Run the adversary $(\tilde{M},\tilde{\markingkey}) \leftarrow \mathcal{A}(\mathcal{O}^f,\hat{M},\verificationkey)$.
		\item $\mathcal{A}$ wins if $\verify(\tilde{\markingkey},\verificationkey,\tilde{M})=1$.
	\end{enumerate}

\end{itemize}

Two other properties, which might be of practical interest but are either too complex to achieve or contrary to our definitions, are \emph{Ownership Piracy} and different degrees of \emph{Verifiability},
\begin{itemize}
	\item \emph{Ownership Piracy} means that an attacker is attempting to implant his watermark into a model which has already been watermarked before. Here, the goal is that the old watermark at least persists. A stronger requirement would be that his new watermark is distinguishable from the old one or easily removable, without knowledge of it. Indeed, we will later show in Section \ref{sec:const:ownershippiracy} that a version of our practical construction fulfills this strong definition. On the other hand, a removable watermark is obviously in general inconsistent with  \emph{Unremovability}, so we leave\footnote{Indeed, Ownership Piracy is only meaningful if the watermark was originally inserted during $\train$, whereas the adversary will have to make adjustments to a pre-trained model. This gap is exactly what we explore in Section \ref{sec:const:ownershippiracy}.} it out in our theoretical construction.
	
	\item A watermarking scheme that uses the verification procedure $\verify$ is called  \emph{privately verifiable}. In such a setting, one can convince a third party about ownership using $\verify$ as long as this third party is honest and does not release the key pair $(\markingkey,\verificationkey)$, which crucially is input to it. We call a scheme \emph{publicly verifiable} if there exists an interactive protocol $\publicverify$ that, on input $\markingkey,\verificationkey,M$ by the prover and $\verificationkey,M$ by the verifier outputs the same value as $\verify$ (except with negligible probability), such that the same key $\verificationkey$ can be used in multiple proofs of ownership.
\end{itemize}

\section{Watermarking From Backdooring}
\label{sec:wmfrombackdooring} 

This section gives a theoretical construction of privately verifiable watermarking based on any strong backdooring (as outlined in 
\secref{preliminaries}) and a commitment scheme.
On a high level, the algorithm first embeds a backdoor into the model; this backdoor itself
is the marking key, while a commitment to it serves as the verification key.

More concretely, let $(\train,\classify)$ be an $\epsilon$-accurate ML algorithm, $\backdoor$ be a strong backdooring algorithm and $(\commit,\open)$ be a statistically hiding commitment scheme. Then define the three algorithms $(\keygen,\marking,\verify)$ as follows.
\ifshort
\paragraph*{$\keygen():$} ~
	\begin{enumerate}
		\item Run $(T,T_L)=\mathsf{b}\sample \samplebd(\mathcal{O}^f)$ where \\ $T=\{t^{(1)},\dots,t^{(n)}\}$ and $T_L=\{T_L^{(1)},\dots,T_L^{(n)} \}$.
		\item Sample $2n$ random strings $r_t^{(i)},r_L^{(i)}\sample \{0,1\}^n$ and generate $2n$ commitments $\{c_{t}^{(i)}, c_L^{(i)}\}_{i \in [n]}$ where $c_t^{(i)}\leftarrow \commit(t^{(i)}, r_t^{(i)})$, $c_L^{(i)}\leftarrow \commit(T_L^{(i)},r_L^{(i)})$.
		\item Set $\markingkey \leftarrow (\mathsf{b}, \{ r_t^{(i)}, r_L^{(i)} \}_{i \in [n]} )$, $\verificationkey \leftarrow \{ c_t^{(i)}, c_L^{(i)} \}_{i \in [n]}$ and return $(\markingkey,\verificationkey)$.
	\end{enumerate}
\paragraph*{$\marking(M,\markingkey):$} ~
	\begin{enumerate}
		\item Let $\markingkey = ( \mathsf{b}, \{ r_t^{(i)}, r_L^{(i)} \}_{i \in [n]} )$.
		\item Compute and output $\hat{M} \leftarrow \backdoor ( \mathcal{O}^f,\mathsf{b}, M)$.
	\end{enumerate}
\paragraph*{$\verify(\markingkey,\verificationkey,M):$} ~
	\begin{enumerate}
		\item\label{step:verify:testifbackdoor} Let  $\markingkey = (\mathsf{b}, \{ r_t^{(i)}, r_L^{(i)} \}_{i \in [n]} )$, $\verificationkey = \{ c_t^{(i)}, c_L^{(i)} \}_{i \in [n]}$. For  $\mathsf{b}=(T,T_L)$ test if $\forall t^{(i)}\in T:~ T_L^{(i)}\neq f(t^{(i)})$. If not, then output $0$.
		\item\label{step:verify:testcommitments} For all $i \in [n]$ check that $\open(c_t^{(i)},t^{(i)},r_t^{(i)})=1$ and \\ $\open(c_L^{(i)},T_L^{(i)},r_L^{(i)})=1$. Otherwise output $0$.
		\item\label{step:verify:testclassify} For all $i\in [n]$ test that $\classify(t^{(i)},M)=T_L^{(i)}$. If this is true for all but $\epsilon |T|$ elements from $T$ then  output $1$, else output $0$.
	\end{enumerate}
\fi

We want to remark that this construction captures both the watermarking of an existing model and the training
from scratch. We now prove the security of the construction.
\begin{thm}
Let $\overline{D}$ be of super-polynomial size in $n$. Then assuming the existence of a commitment scheme  and a strong backdooring scheme, the aforementioned algorithms $(\keygen,\marking,\verify)$ form a privately verifiable watermarking scheme.
\end{thm}

The proof, on a very high level, works as follows: a model containing a strong backdoor means that this backdoor, and therefore the watermark, cannot be removed. Additionally, by the hiding property of the commitment scheme the verification key will not provide any useful information to the adversary about the backdoor used, while the binding property ensures that one cannot claim ownership of arbitrary models.
In the proof, special care must be taken as we use reductions from the watermarking algorithm to the security of both the underlying backdoor and the commitment scheme. To be meaningful, those reductions must have much smaller runtime than actually breaking these assumptions directly. While this is easy in the case of the commitment scheme, reductions to backdoor security need more attention.

\begin{proof} We prove the following properties:
	\paragraph{Correctness.} 
By construction, $\hat{M}$ which is returned by $\marking$ will disagree with $\mathsf{b}$ on elements from $ T$ with probability at most  $\epsilon$, so in total at least $(1 -\epsilon) |T|$ elements agree by the definition of a backdoor. $\verify$ outputs $1$ if $\hat{M}$ disagrees with $\mathsf{b}$ on at most $\epsilon |T|$ elements. 
	
	\paragraph{Functionality-preserving.}
	Assume that $\backdoor$ is a backdooring algorithm, then by its definition the model $\hat{M}$ is accurate outside of the trigger set of the backdoor, i.e. 
	\begin{align*}
		\Pr_{x \in \overline{D} \setminus T } \left[ f(x)\neq \classify(\hat{M},x) \right] \leq \epsilon\text{.} 
	\end{align*}
	$\hat{M}$ in total will then err on a fraction at most $\epsilon'=\epsilon + n/|D|$, and because $\overline{D}$ by assumption is super-polynomially large in $n$ $\epsilon'$ is negligibly close to $\epsilon$.

	\paragraph{Non-trivial ownership.} To win, $\mathcal{A}$ must guess the correct labels for a $1-\epsilon$ fraction of $\tilde{T}$ in advance,
as $\mathcal{A}$ cannot change the chosen value $\tilde{T},\tilde{T_L}$ after seeing the model due to the binding property of the commitment scheme. As $\keygen$ chooses the set $T$ in $\markingkey$ uniformly at random, whichever set $\mathcal{A}$ fixes for $\tilde{\markingkey}$ will intersect with $T$ only with negligible probability by definition (due to the \emph{multiple trigger sets} property). So assume for simplicity that $\tilde{T}$ does not intersect with $T$.
	Now $\mathcal{A}$ can choose $\tilde{T}$ to be of elements either from within $\overline{D}$ or outside of it. Let $n_1=|\overline{D}\cap \tilde{T}|$ and $n_2=|\tilde{T}|-n_1$. 
	
	For the benefit of the adversary, we make the strong assumption that whenever $M$ is inaccurate for $x\in \overline{D}\cap \tilde{T}$ then it classifies to the label in $\tilde{T}_L$. But as $M$ is $\epsilon$-accurate on $\overline{D}$, the ratio of incorrectly classified committed labels is $(1-\epsilon)n_1$. 
For every choice $\epsilon<0.5$ we  have that $\epsilon n_1 < (1-\epsilon)n_1$. Observe that for our scheme, the value $\epsilon$ would be chosen much smaller than $0.5$ and therefore this inequality  always holds.
	
	On the other hand, let's look at  all values of $\tilde{T}$ that lie in $D\setminus \overline{D}$. By the assumption about machine learning that we made in its definition, if the input was chosen independently of $M$ and it lies outside of $\overline{D}$ then  $M$ will in expectancy misclassify  $\frac{|L|-1}{|L|}n_2$ elements. We then have that $\epsilon n_2 < \frac{|L|-1}{|L|}n_2$ as $\epsilon <0.5$ and $L\geq 2$. As $\epsilon n = \epsilon n_1 + \epsilon n_2$, the error of $\tilde{T}$ must be larger than $\epsilon n$.
	
	\paragraph{Unremovability.}
Assume that there exists no algorithm that can generate an $\epsilon$-accurate model $N$ in time $t$ of $f$, where $t$ is a lot smaller that the time necessary for training such an accurate model using $\train$. At the same time, assume that the adversary $\mathcal{A}$ breaking the unremovability property takes time approximately $t$.
By definition, after running $\mathcal{A}$ on input $M,\verificationkey$ it will output a model $\tilde{M}$ which will be $\epsilon$-accurate and at least a $(1-\epsilon)$-fraction of the elements from the set $T$ will be classified correctly. The goal in the proof is to show that $\mathcal{A}$ achieves this independently of $\verificationkey$.	
	In a first step, we will use a hybrid argument to show that $\mathcal{A}$ essentially works independent of  $\verificationkey$. Therefore, we construct a series of algorithms where we gradually replace the backdoor elements in $\verificationkey$. First, consider the following algorithm $\mathcal{S}$:
		\begin{enumerate}
		\item  Compute $(M,\hat{M},\markingkey,\verificationkey) \leftarrow \MModel()$.
		\item Sample $(\tilde{T},\tilde{T}_L) = \tilde{\mathsf{b}} \sample \samplebd(\mathcal{O}^f)$ where  $\tilde{T}=\{\tilde{t}^{(1)},\dots, \tilde{t}^{(n)}\}$ and $\tilde{T}_L=\{\tilde{T}_L^{(1)},\dots,\tilde{T}_L^{(n)} \}$. Now set
\begin{equation*}
c_t^{(1)}\leftarrow \commit(\tilde{t}^{(1)}, r_t^{(1)}),  c_L^{(1)}\leftarrow \commit(\tilde{T}_L^{(1)},r_L^{(1)})
\end{equation*} 
and  $ \tilde{\verificationkey} \leftarrow \{ c_t^{(i)}, c_L^{(i)} \}_{i \in [n]}$

	\item Compute $\tilde{M} \leftarrow \mathcal{A}(\mathcal{O}^f,\hat{M},\tilde{\verificationkey})$.
\end{enumerate}
This algorithm replaces the first element in a verification key with an element from an independently generated backdoor, and then runs $\mathcal{A}$ on it.

In $\mathcal{S}$ we only exchange one commitment when compared to the input distribution to $\mathcal{A}$ from the security game. By the statistical hiding of $\commit$, the output of $\mathcal{S}$ must be distributed statistically close to the output of $\mathcal{A}$ in the unremovability experiment. 
Applying this repeatedly, we construct a sequence of hybrids $\mathcal{S}^{(1)},\mathcal{S}^{(2)},\dots,\mathcal{S}^{(n)}$ that change $1,2,\dots,n$ of the elements from $\verificationkey$ in the same way that $\mathcal{S}$ does and conclude that the success of outputting a model $\tilde{M}$ without the watermark using $\mathcal{A}$ must be independent of $\verificationkey$.

Consider the following algorithm $\mathcal{T}$ when given a model $M$ with a strong backdoor: 
\begin{enumerate}
		\item Compute $(\markingkey,\verificationkey)\leftarrow \keygen()$.
	\item Run the adversary and compute $\tilde{N} \leftarrow \mathcal{A}(\mathcal{O}^f,M,\verificationkey)$. 
\end{enumerate}
By the hybrid argument above, the algorithm $\mathcal{T}$ runs nearly in the same time as $\mathcal{A}$, namely $t$, and its output $\tilde{N}$ will be without the backdoor that $M$ contained. But then, by persistence of strong backdooring, $\mathcal{T}$ must also generate $\epsilon$-accurate models given arbitrary, in particular bad input models $M$ in the same time $t$, which contradicts our assumption that no such algorithm exists.
	
\paragraph{Unforgeability.}
Assume that there exists a poly-time algorithm $\mathcal{A}$ that can break unforgeability. We will use this algorithm to open a statistically hiding commitment.

Therefore, we design an algorithm $\mathcal{S}$ which uses $\mathcal{A}$ as a subroutine. The algorithm trains a regular network (which can be watermarked by our scheme) and adds the commitment  into the verification key. Then, it will use $\mathcal{A}$ to find openings for these commitments. The algorithm $\mathcal{S}$ works as follows:
\begin{enumerate}
	\item Receive the commitment $c $ from challenger.
	\item Compute $(M,\hat{M},\markingkey,\verificationkey) \leftarrow \MModel()$.
	\item Let $\verificationkey = \{ c_t^{(i)}, c_L^{(i)} \}_{i \in [n]}$  set 
	\[
	 \hat{c}_t^{(i)} \leftarrow \begin{cases}
	 c & \text{ if } i=1 \\
	 c_t^{(i)} & \text{ else}
	 \end{cases}
	\]
	and $\hat{\verificationkey}\leftarrow \{ \hat{c}_t^{(i)},c_L^{(i)} \}_{i\in [n]}$.
	\item Compute $(\tilde{M},\tilde{\markingkey})\leftarrow \mathcal{A}(\mathcal{O}^f,\hat{M},\hat{\verificationkey})$.
	\item Let $\tilde{\markingkey} = ( (\{t^{(1)},\dots, t^{(n)}\},T_L),\{ r_t^{(i)}, r_L^{(i)} \}_{i \in [n]} )$.
	
	 If $\verify(\tilde{\markingkey},\hat{\verificationkey},\tilde{M})=1$ output $ t^{(1)},r_t^{(1)}$,  else output $\bot$.
\end{enumerate}
Since the commitment scheme is statistically hiding, the input to $\mathcal{A}$ is statistically indistinguishable from an input where $\hat{M}$ is backdoored on all the committed values of $\verificationkey$. Therefore the output of $\mathcal{A}$ in $\mathcal{S}$ is statistically indistinguishable from the output in the unforgeability definition. With the same probability as in the definition, $\tilde{\markingkey},\hat{\verificationkey},\tilde{M}$ will make $\verify$ output $1$. But by its definition, this means that $\open(c,t^{(1)},r_t^{(1)})=1$ so $t^{(1)},r_t^{(1)}$ open the challenge commitment $c$. As the commitment is statistically hiding (and we generate the backdoor independently of $c$) this will open $c$ to another value then for which it was generated with overwhelming probability.
\end{proof}

%

\subsection{From Private to Public Verifiability}
\label{sec:pubver}
Using the algorithm $\verify$ constructed in this section only allows verification by an honest party. The scheme described above is therefore only privately verifiable. After running $\verify$, the key $\markingkey$ will be known and an adversary can retrain the model  on the trigger set. This is not a drawback when it comes to an application like the protection of intellectual property, where a trusted third party in the form of a judge exists.  If one instead wants to achieve public verifiability, then there are two possible scenarios for how to design an algorithm $\publicverify$: allowing public verification a constant number of times, or an arbitrary number of times.

\begin{figure}[h]
	\begin{center}
		\includegraphics[width=22em,height=9em]{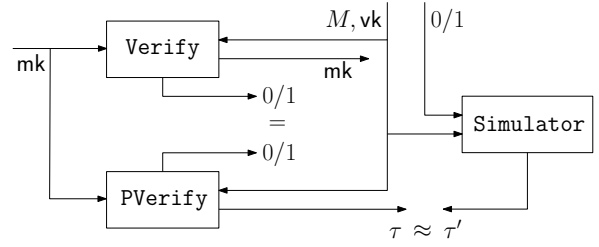}
	\end{center}
	\caption{A schematic illustration of the public verification process.}
	\label{fig:simulation}
\end{figure}

In the first setting, a straightforward approach to the construction of $\publicverify$ is to choose multiple backdoors during $\keygen$ and release a different one in each iteration of $\publicverify$. This allows multiple verifications, but the number is upper-bounded in practice by the capacity of the model $M$ to contain backdoors - this cannot arbitrarily be extended without damaging the accuracy of the model. To achieve an unlimited number of verifications we will modify the watermarking scheme to output a different type of verification key. We then present an algorithm $\publicverify$ such that the interaction $\tau$ with an honest prover can be simulated as $\tau'$ given the values $M,\verificationkey,\verify(\markingkey,\verificationkey,M)$ only. This simulation means that no other information about $\markingkey$ beyond what is leaked from $\verificationkey$ ever gets to the verifier. We give a graphical depiction of the approach in Figure \ref{fig:simulation}. Our solution is sketched in Appendix \ref{app:publicver}.

\subsection{Implementation Details}
For an implementation, it is of importance to choose the size $|T|$ of the trigger set properly, where we have to consider that $|T|$ cannot be arbitrarily big, as the accuracy will drop. To lower-bound $|T|$ we assume an attacker against non-trivial ownership. For simplicity, we use a backdooring algorithm that generates trigger sets from elements where $f$ is undefined. By our simplifying assumption from Section \ref{subsec:ml}, the model will classify the images in the trigger set to random labels. Furthermore, assume that the model is $\epsilon$-accurate (which it also is on the trigger set). Then, one can model a dishonest party to randomly get $(1-\epsilon)|T|$ out of $|T|$ committed images right using a Binomial distribution. We want to upper-bound this event to have probability at most $2^{-n}$ and use Hoeffding's inequality to obtain that $|T|>n\cdot \ln(2)/(\frac{1}{|L|} +\epsilon -1)$.  

To implement our scheme, it is necessary that $\verificationkey$ becomes public before $\verify$ is used. This ensures that a party does not simply generate a fake key after seeing a model. A solution for this is to e.g. publish the key on a time-stamped bulletin board like a blockchain. 
In addition, a statistically hiding commitment scheme should be used that allows for efficient evaluation in zero-knowledge (see Appendix \ref{app:publicver}). For this one can e.g. use a scheme based on a cryptographic hash function such as the one described in \cite{nigel_cryptomadesimple}.

\section{A Direct Construction of Watermarking} 
\seclab{experiments}

This section describes a scheme for watermarking a neural network model for image classification, and  experiments analyzing it with respect to the definitions in Section \ref{sec:model}. We demonstrate that it is hard to reduce the persistence of watermarks that are generated with our method. For all the technical details regarding the implementation and hyper-parameters, we refer the reader to Section~\ref{sec:techdet}.

\subsection{The Construction}
Similar to Section \ref{sec:wmfrombackdooring}, we use a set of images as the \emph{marking key} or \emph{trigger set} of our construction\footnote{As the set of images will serve a similar purpose as the trigger set from backdoors in \secref{preliminaries}, we denote the marking key as trigger set throughout this section.}. To embed the watermark, we optimize the models using both training set and trigger set. We investigate two approaches: the first approach starts from a pre-trained model, i.e., a model that was trained without a trigger set, and continues training the model together with a chosen trigger set. This approach is denoted as \pretrain. The second approach trains the model from scratch along with the trigger set. This approach is denoted as \scratch. This latter approach is related to \emph{Data Poisoning} techniques.

During training, for each batch, denote as $b_t$ the batch at iteration $t$,  we sample $k$ trigger set images and append them to $b_t$. We follow this procedure for both approaches. We tested different numbers of $k$ (i.e., 2, 4, and 8), and setting $k=2$ reach the best results. We hypothesize that this is due to the \emph{Batch-Normalization} layer~\cite{ioffe2015batch}. The Batch-Normalization layer has two modes of operations. During training, it keeps a running estimate of the computed mean and variance. During an evaluation, the running mean and variance are used for normalization. Hence, adding more images to each batch puts more focus on the trigger set images and makes convergence slower.

In all models we optimize the Negative Log Likelihood loss function on both training set and \emph{trigger set}. Notice, we assume the creator of the model will be the one who embeds the watermark, hence has access to the training set, test set, and \emph{trigger set}.

In the following subsections, we demonstrate the efficiency of our method regarding non-trivial ownership and unremovability and furthermore show that it is functionality-preserving, following the ideas outlined in Section \ref{sec:model}. For that we use three different image classification datasets: CIFAR-10, CIFAR-100 and ImageNet \cite{krizhevsky2009learning, russakovsky2015imagenet}. We chose those datasets to demonstrate that our method can be applied to models with a different number of classes and also for large-scale datasets.

\subsection{Non-Trivial Ownership}
In the \emph{non-trivial ownership} setting, an adversary will not be able to claim ownership of the model even if he knows the watermarking algorithm. To fulfill this requirement we randomly sample the examples for the trigger set. We sampled a set of 100 abstract images, and for each image, we randomly selected a target class.

\begin{figure}[h!]
  \centering
  \includegraphics[width=0.42\textwidth]{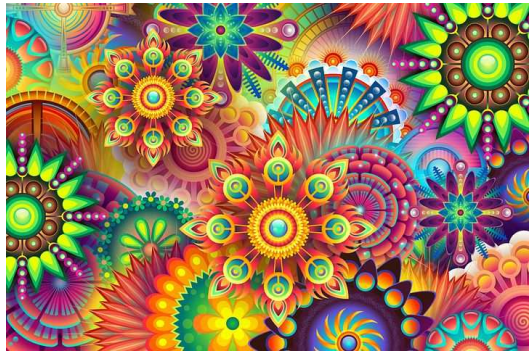}
  \caption{An example image from the trigger set. The label that was assigned to this image was ``automobile''.}
  \label{fig:wm_exm}
\end{figure}

This sampling-based approach ensures that the examples from the trigger set are uncorrelated to each other. Therefore revealing a subset from the trigger set will not reveal any additional information about the other examples in the set, as is required for public verifiability. Moreover, since both examples and labels are chosen randomly, following this method makes back-propagation based attacks extremely hard. Figure~\ref{fig:wm_exm} shows an example from the trigger set.

\subsection{Functionality-Preserving}
For the \emph{functionality-preserving} property we require that a model with a watermark should be as
accurate as a model without a watermark. In general, each task defines its own measure of performance~\cite{adi2016structed,keshet2014optimizing,adi2016automatic,adi2017sequence}. However, since in the current work we are focused on
image classification tasks, we measure the accuracy of the model using the 0-1 loss.

Table~\ref{tb:func_pres_res} summarizes the test set and trigger-set classification accuracy on CIFAR-10 and CIFAR-100, for three different models; (i) a model with no watermark (\nowm); (ii) a model that was trained with the trigger set from scratch (\scratch); and (iii) a pre-trained model that was trained with the trigger set after convergence on the original training data set (\pretrain).

\begin{table} [h!]
  \centering  
  \begin{tabular}{ l | p{1.8cm} | p{1.8cm} }
  
     Model & Test-set acc. & Trigger-set acc. \\ 
    \hline        
    \hline
    \multicolumn{3}{c}{CIFAR-10}\\
    \hline
    \nowm & 93.42 & 7.0 \\ \hline
    \scratch & 93.81 & 100.0 \\ \hline
    \pretrain & 93.65 & 100.0 \\ \hline
    \multicolumn{3}{c}{CIFAR-100}\\
    \hline
    \nowm & 74.01 & 1.0 \\ \hline
    \scratch & 73.67 & 100.0 \\ \hline
    \pretrain & 73.62 & 100.0 \\ \hline
    \hline
  \end{tabular}
  \caption{Classification accuracy for CIFAR-10 and CIFAR-100 datasets on the test set and trigger set.}
  \label{tb:func_pres_res}
\end{table}

It can be seen that all models have roughly the same test set accuracy and that in both  \scratch~and \pretrain~the trigger-set accuracy is 100\%. Since the trigger-set labels were chosen randomly, the \nowm~models' accuracy depends on the number of classes. For example, the accuracy on CIFAR-10 is 7.0\% while on CIFAR-100 is only 1.0\%.

\subsection{Unremovability}

In order to satisfy the \emph{unremovability} property, we first need to define the types of unremovability
functions we are going to explore. Recall that our goal in the unremovability experiments is to investigate
the robustness of the watermarked models against changes that aim to remove the watermark while keeping the
same functionality of the model. Otherwise, one can set all  weights to zero and completely remove the watermark but also destroy the model. 

Thus, we are focused on \emph{fine-tuning} experiments. In other words, we wish to keep or improve the
performance of the model on the test set by carefully training it. Fine-tuning seems to be the most probable
type of attack since it is frequently used and requires less computational resources and training
data~\cite{simonyan2014very, yosinski2014transferable, razavian2014cnn}. Since in our settings we would like
to explore the robustness of the watermark against strong attackers, we assumed that the adversary can
fine-tune the models using the same amount of training instances and epochs as in training the model.

An important question one can ask is: \emph{when is it still my model?} or other words how much can I change the model and still claim ownership? This question is highly relevant in the case of watermarking. In the current work we handle this issue by measuring the performance of the model on the test set and trigger set, meaning that the original creator of the model can claim ownership of the model if the model is still $\epsilon$-accurate on the original test set while also $\epsilon$-accurate on the trigger set. We leave the exploration of different methods and of a theoretical definition of this question for future work.

\paragraph{Fine-Tuning.}
We define four different variations of fine-tuning procedures: 
\begin{itemize}
	\item \emph{Fine-Tune Last Layer} (FTLL): Update the parameters of the last layer only. In this setting we freeze the parameters in all the layers except in the output layer. One can think of this setting as if the model outputs a new representation of the input features and we fine-tune only the output layer.
    \item \emph{Fine-Tune All Layers} (FTAL): Update all the layers of the model.
    \item \emph{Re-Train Last Layers} (RTLL): Initialize the parameters of the output layer with random weights and only update them. In this setting, we freeze the parameters in all the layers except for the output layer. The motivation behind this approach is to investigate the robustness of the watermarked model under noisy conditions. This can alternatively be seen as changing the model to classify for a different set of output labels.
    \item \emph{Re-Train All Layers} (RTAL): Initialize the parameters of the output layer with random weights and update the parameters in all the layers of the network.
\end{itemize}

\begin{figure}[h!]
  \includegraphics[width=\linewidth]{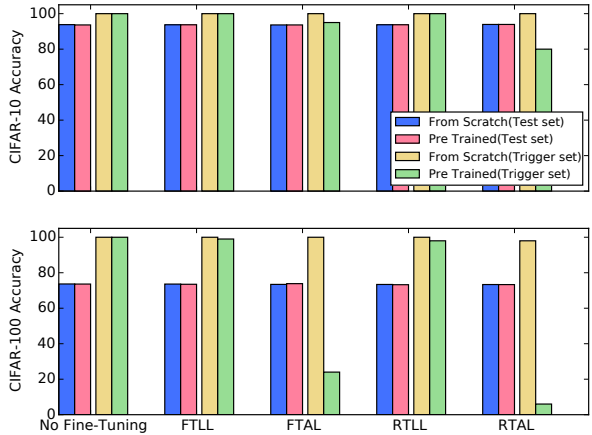}
  \caption{Classification accuracy on the test set and trigger set for CIFAR-10 (top) and CIFAR-100 (bottom)
    using different fine-tuning techniques. For example, in the bottom right bars we can see that the \pretrain~model (green) suffers a dramatic decrease in the results comparing the baseline (bottom left) using the RTAL technique.}
  \label{fig:cifar_tinetune}
\end{figure}

Figure~\ref{fig:cifar_tinetune} presents the results for both the \pretrain~and \scratch~models over the test set and trigger set, after applying these four different fine-tuning techniques.

The results suggest that while both models reach almost the same accuracy on the test set, the \scratch~models are superior or equal to the \pretrain~models overall fine-tuning methods. \scratch~reaches roughly the same accuracy on the trigger set when each of the four types of fine-tuning approaches is applied.

Notice that this observation holds for both the CIFAR-10 and CIFAR-100 datasets, where for CIFAR-100 it appears to be easier to remove the trigger set using the \pretrain~models. Concerning the above-mentioned results, we now investigate what will happen if an adversary wants to embed a watermark in a model which has already been watermarked. This can be seen as a black-box attack on the already existing watermark. According to the fine-tuning experiments, removing this new trigger set using the above fine-tuning approaches will not hurt the original trigger set and will dramatically decrease the results on the new trigger set. In the next paragraph, we explore and analyze this setting. Due to the fact that \scratch~models are more robust than \pretrain, for the rest of the paper, we report the results for those models only.



\subsection{Ownership Piracy}\label{sec:const:ownershippiracy}
As we mentioned in Section \ref{sec:model}, in this set of experiments we explore the scenario where an adversary wishes to claim ownership of a model which has already been watermarked. 

For that purpose, we collected a new trigger set of different 100 images, denoted as \tsnew, and embedded it to the \scratch~model (this new set will be used by the adversary to claim ownership of the model). Notice that the \scratch~models were trained using a different trigger set, denoted as \tsorig. Then, we fine-tuned the models using RTLL and RTAL methods. In order to have a fair comparison between the robustness of the trigger sets after fine-tuning, we use the same amount of epochs to embed the new trigger set as we used for the original one.

\begin{figure}[h!]
  \includegraphics[width=\linewidth]{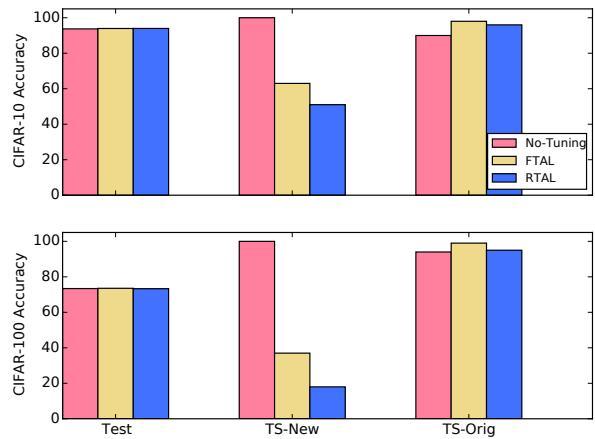}
  \caption{Classification accuracy on CIFAR-10 (top) and CIFAR-100 (bottom) datasets after embedding two trigger sets, \tsorig~and \tsnew. We present results for no tuning (red), FTAL (yellow) and TRAL (blue).}
  \label{fig:cifar_2wms}
\end{figure}

Figure~\ref{fig:cifar_2wms} summarizes the results on the test set, \tsnew~and \tsorig. We report results for both the FTAL and RTAL methods together with the baseline results of no fine tuning at all (we did not report here the results of FTLL and RTLL since those can be considered as the easy cases in our setting). The red bars refer to the model with no fine tuning, the yellow bars refer to the FTAL method and the blue bars refer to RTAL.

The results suggest that the original trigger set, \tsorig, is still embedded in the model (as is demonstrated in the right columns) and that the accuracy of classifying it even improves after fine-tuning. This may imply that the model embeds the trigger set in a way that is close to the training data distribution. However, in the new trigger set, \tsnew, we see a significant drop in the accuracy. Notice, we can consider embedding \tsnew~as embedding a watermark using the \pretrain~approach. Hence, this accuracy drop of \tsnew~is not surprising and goes in hand with the results we observed in Figure~\ref{fig:cifar_tinetune}.


%


\paragraph{Transfer Learning.}
In transfer learning we would like to use knowledge gained while solving one problem and apply it to a different problem. For example, we use a trained model on one dataset (source dataset) and fine-tune it on a new dataset (target dataset). For that purpose, we fine-tuned the \scratch~model (which was trained on either CIFAR-10 or CIFAR-100), for another 20 epochs using the labeled part of the STL-10 dataset~\cite{coates2011analysis}. 

Recall that our watermarking scheme is based on the outputs of the model. As a result, when fine-tuning a model on a different dataset it is very likely that we change the number of classes, and then our method will probably break. Therefore, in order to  still be able to verify the watermark we save the original output layer, so that on verification time we use the model's original output layer instead of the new one.

Following this approach makes both FTLL and RTLL useless due to the fact that these methods update the parameters of the output layer only. Regarding FTAL, this approach makes sense in specific settings where the classes of the source dataset are related to the target dataset. This property holds for CIFAR-10 but not for CIFAR-100. Therefore  we report the results only for RTAL method. 

Table~\ref{tb:cifarrestransfer} summarizes the classification accuracy on the test set of STL-10 and the trigger set after transferring from CIFAR-10 and CIFAR-100.

\begin{table} [h!]
  \centering
  \scalebox{0.95}{
  \begin{tabular}{ l | c | c }
     & Test set acc. & Trigger set acc.\\ 
    \hline
    \hline
    CIFAR10 $\rightarrow$ STL10 & 81.87 & 72.0\\ \hline
    CIFAR100 $\rightarrow$ STL10 & 77.3 & 62.0\\ \hline
    \hline
  \end{tabular}}
\caption{Classification accuracy on STL-10 dataset and the trigger set, after transferring from either CIFAR-10 or CIFAR-100 models.}
   \label{tb:cifarrestransfer}
\end{table}

Although the trigger set accuracy is smaller after transferring the model to a different dataset, results suggest that the trigger set still has a lot of presence in the network even after fine-tuning on a new dataset.



\subsection{ImageNet - Large Scale Visual Recognition Dataset}

For the last set of experiments, we would like to explore the robustness of our watermarking method on a large scale dataset. For that purpose, we use ImageNet dataset~\cite{russakovsky2015imagenet} which contains about 1.3 million training images with over 1000 categories.

Table~\ref{tb:imagenetres} summarizes the results for the \emph{functionality-preserving} tests. We can see from Table~\ref{tb:imagenetres} that both models, with and without watermark, achieve roughly the same accuracy in terms of Prec@1 and Prec@5,  while the model without the watermark attains 0\% on the trigger set and the watermarked model attain 100\% on the same set.

\begin{table} [h!]
  \centering
  \scalebox{0.95}{
  \begin{tabular}{ l | c | c}
     & Prec@1 & Prec@5  \\ 
    \hline  
    \hline
    \multicolumn{3}{c}{Test Set}\\
    \hline
    \nowm  & 66.64 & 87.11 \\ \hline    
    \scratch & 66.51 & 87.21\\ \hline
    \hline  
    \multicolumn{3}{c}{Trigger Set}\\
    \hline
    \nowm  & 0.0 & 0.0 \\ \hline    
    \scratch & 100.0 & 100.0 \\ \hline
    \hline
  \end{tabular}}
  \caption{ImageNet results, Prec@1 and Prec@5, for a ResNet18 model with and without a watermark.}
  \label{tb:imagenetres}
\end{table}

Notice that the results we report for ResNet18 on ImageNet are slightly below what is reported in the literature. The reason beyond that is due to training for fewer epochs (training a model on ImageNet is computationally expensive, so we train our models for fewer epochs than what is reported). 

In Table~\ref{tb:imagenettransfer} we report the results of transfer learning from ImageNet to ImageNet, those can be considered as FTAL, and from ImageNet to CIFAR-10, can be considered as RTAL or transfer learning.

\begin{table} [h!]
  \centering
  \scalebox{0.95}{
  \begin{tabular}{ l | c | c}
     & Prec@1 & Prec@5  \\ 
    \hline  
    \hline
    \multicolumn{3}{c}{Test Set}\\
    \hline
    ImageNet $\rightarrow$ ImageNet & 66.62 & 87.22\\ \hline
    ImageNet $\rightarrow$ CIFAR-10 & 90.53 & 99.77\\ \hline
    \hline  
    \multicolumn{3}{c}{Trigger Set}\\
    \hline
    ImageNet $\rightarrow$ ImageNet & 100.0 & 100.0\\ \hline    
    ImageNet $\rightarrow$ CIFAR-10 & 24.0 & 52.0\\ \hline
    \hline
  \end{tabular}}
  \caption{ImageNet results, Prec@1 and Prec@5, for fine tuning using ImageNet and CIFAR-10 datasets.}
  \label{tb:imagenettransfer}
\end{table}

Notice that after fine tuning on ImageNet, trigger set results are still very high, meaning that the trigger set has a very strong presence in the model also after fine-tuning. When transferring to CIFAR-10, we see a drop in the Prec@1 and Prec@5. However, considering the fact that ImageNet contains 1000 target classes, these results are still significant.


\subsection{Technical Details}
\label{sec:techdet}
We implemented all  models using the PyTorch package \cite{paszke2017automatic}. In all the experiments we used a ResNet-18 model, which is a convolutional based neural network model with 18
layers~\cite{he2016deep, he2016identity}. We optimized each of the models using Stochastic Gradient Descent
(SGD), using a learning rate of 0.1. For CIFAR-10 and CIFAR-100 we trained the models for 60 epochs while halving
the learning rate by ten every 20 epochs. For ImageNet we trained the models for 30 epochs while halving the
learning rate by ten every ten epochs. The batch size was set to 100 for the CIFAR10 and CIFAR100, and to 256 for
ImageNet. For the fine-tuning tasks, we used the last learning rate that was used during training.

\section{Conclusion and Future Work}
In this work we proposed a practical analysis of the ability to watermark a neural network using random training instances and random labels. We presented possible attacks that are both black-box and grey-box in the model, and showed how robust our watermarking approach is to them. At the same time, we outlined a theoretical connection to the previous work on backdooring such models. 

For future work we would like to define a theoretical boundary for how much change must a party apply to a model before he can claim ownership of the model. We also leave as an open problem the construction of a practically efficient zero-knowledge proof for our publicly verifiable watermarking construction.

\section*{Acknowledgments}
This work was supported by the BIU Center for Research in Applied Cryptography and Cyber Security in conjunction with the Israel National Cyber Directorate in the Prime Minister's Office.

{\footnotesize \bibliographystyle{acm}
\bibliography{refs}}

\newpage
\appendix
\section{Supplementary Material}
In this appendix we further discuss how to achieve public verifiability for a variant of our watermarking scheme. Let us first introduce the following additional notation: for a vector $\mathbf{e}\in \{0,1\}^\ell$, let $\mathbf{e}|_0=\{i\in [\ell] ~|~ \mathbf{e}[i]=0  \}$ be the set of all indices where $\mathbf{ e}$ is $0$ and define $\mathbf{ e}|_{1}$ accordingly. Given a verification key $\verificationkey = \{ c_t^{(i)}, c_L^{(i)} \}_{i \in [\ell]}$ containing $\ell$ elements and a vector $\mathbf{e}\in \{0,1\}^\ell$, we write the selection of elements from $\verificationkey$ according to $\mathbf{ e}$ as
\[\verificationkey|_0^{\mathbf{e}}=\{ c_t^{(i)}, c_L^{(i)}  \}_{i\in \mathbf{e}|_0} \quad \text{ and } \quad \verificationkey|_1^{\mathbf{e}} = \{ c_t^{(i)}, c_L^{(i)}  \}_{i\in \mathbf{e}|_1}\text{.} \] 

For a marking key $\markingkey=( \mathsf{b}, \{ r_t^{(i)}, r_L^{(i)} \}_{i \in [\ell]} )$ with $\ell$ elements and  $\mathsf{b}=\{ T^{(i)},T_L^{(i)}  \}_{i \in [\ell]}$  we then define
\[
 \markingkey|_{0}^{\mathbf{e}}=( \mathsf{b}|_0^{\mathbf{e}}, \{ r_t^{(i)}, r_L^{(i)} \}_{i \in \mathbf{e}|_0} )  \text{ with }  \mathsf{b}|_0^{\mathbf{e}}=\{ T^{(i)},T_L^{(i)}  \}_{i \in \mathbf{ e}|_0}
 \]
  (and $\markingkey|_1^{\mathbf{ e}}$ accordingly). We assume the existence of a cryptographic hash function $H:\{0,1\}^{p(n)}\rightarrow \{0,1\}^n$.

\subsection{From Private to Public Verifiability}
\label{app:publicver}

To achieve public verifiability, we will make use of a cryptographic tool called a \emph{zero-knowledge argument} \cite{gmr_zk}, which is a technique that allows a prover $\mathcal{P}$ to convince a verifier $\mathcal{V}$ that a certain public statement is true, without giving away any further information. This idea is similar to the idea of unlimited public verification as outlined in Section \ref{sec:pubver}.

\paragraph{Zero-Knowledge Arguments.} 
Let TM be an abbreviation for Turing Machines. An iTM is defined to be an interactive TM, i.e. a Turing Machine with a special communication tape. Let $L_R\subseteq \{0,1\}^*$ be an NP language and $R$ be its related NP-relation, i.e. $(x,w)\in R$ iff $x\in L_R$ and the TM used to define $L_R$ outputs $1$ on input of the statement $x$ and the witness $w$. We write $R_x=\{w ~| ~ (x,w)\in R\}$ for the set of witnesses for a fixed $x$. Moreover, let $\mathcal{P},\mathcal{V}$ be a pair of PPT iTMs. For $(x,w)\in R$, $\mathcal{P}$ will obtain $w$ as input while $\mathcal{V}$ obtains an auxiliary random string $z\in \{0,1\}^*$. In addition, $x$ will be input to both TMs. Denote with $\mathcal{V}^{\mathcal{P}(a)}(b)$ the output of the iTM $\mathcal{V}$ with input $b$ when communicating with an instance of $\mathcal{P}$ that has input $a$.

$(\mathcal{P},\mathcal{V})$ is called an \emph{interactive proof system} for the language $L$ if the following two conditions hold:
\begin{description}
	\item[Completeness:] For every $x\in L_R$ there exists a string $w$ such that for every $z$: $\Pr[ \mathcal{V}^{\mathcal{P}(x,w)}(x,z)=1 ]$ is negligibly close to $1$.
	\item[Soundness:] For every $x \not\in L_R$, every PPT iTM $\mathcal{P}^{*}$ and every string $w,z$: $\Pr[ \mathcal{V}^{\mathcal{P}^*(x,w)}(x,z)=1 ]$ is negligible.
\end{description}

An interactive proof system is called computational \emph{zero-knowledge} if for every PPT $\hat{\mathcal{V}}$ there exists a PPT simulator $\mathcal{S}$ such that for any $x\in L_R$
\[ \{ \hat{V}^{\mathcal{P}(x,w)}(x,z) \}_{ w \in R_x ,z\in \{0,1\}^*} \approx_c  \{ \mathcal{S}(x,z) \}_{z \in \{0,1\}^*}\text{,}
\]
meaning that all information which can be learned from observing a protocol transcript can also be obtained from running a polynomial-time simulator $\mathcal{S}$ which has no knowledge of the witness $w$.
\subsubsection{Outlining the Idea}
An intuitive approach to build $\publicverify$ is to convert the algorithm $\verify(\markingkey,\verificationkey,M)$ from  Section \ref{sec:wmfrombackdooring} into an NP relation $R$ and use a zero-knowledge argument system. Unfortunately, this must fail due to Step \ref{step:verify:testifbackdoor} of $\verify$: there, one tests if the item $\mathsf{b}$ contained in $\markingkey$ actually is a backdoor as defined above. Therefore, we would need access to the ground-truth function $f$ in the interactive argument system. This first of all needs human assistance, but is moreover only possible by revealing the backdoor elements.

We will now give a different version of the scheme from Section \ref{sec:wmfrombackdooring} which embeds an additional proof into $\verificationkey$. This proof shows that, with overwhelming probability, most of the elements in the verification key indeed form a backdoor. Based on this, we will then design a different verification procedure, based on a zero-knowledge argument system.

\subsubsection{A Convincing Argument that most Committed Values are Wrongly Classified}
Verifying that most of the elements of the trigger set are labeled wrongly is possible, if one accepts\footnote{This is fine if $T$, as in our experiments, only consists of random images.} to release a portion of this set.  To solve the proof-of-misclassification problem, we use the so-called  \emph{cut-and-choose} technique: in cut-and-choose, the verifier $\mathcal{V}$ will ask the prover $\mathcal{P}$ to open a subset of the committed inputs and labels from the verification key. Here, $\mathcal{V}$ is allowed to choose the subset that will be opened to him. Intuitively, if $\mathcal{P}$  committed to a large number elements that are correctly labeled (according to $\mathcal{O}_f$), then at least one of them will show up in the values opened by $\mathcal{P}$ with overwhelming probability over the choice that $\mathcal{V}$ makes. Hence, most of the remaining commitments which were not opened must form a correct backdoor.

To use cut-and-choose, the backdoor size must contain $\ell> n$ elements, where our analysis will use $\ell=4n$ (other values of $\ell$ are also possible).  Then, consider the following protocol  between  $\mathcal{P}$ and $\mathcal{V}$:
\paragraph*{$\cnc(\ell):$} ~
\begin{enumerate}
	\item $\mathcal{P}$ runs $(\markingkey,\verificationkey)\leftarrow \keygen(\ell)$ to obtain a backdoor of size $\ell$ and sends $\verificationkey$ to $\mathcal{V}$. We again define $\markingkey = (\mathsf{b}, \{ r_t^{(i)}, r_L^{(i)} \}_{i \in [\ell]} )$, $\verificationkey = \{ c_t^{(i)}, c_L^{(i)} \}_{i \in [\ell]}$
	\item $\mathcal{V}$ chooses  $\mathbf{e}\leftarrow \{0,1\}^{\ell}$ uniformly at random and sends it to $\mathcal{P}$.
	\item $\mathcal{P}$ sends $\markingkey|_1^{\mathbf{ e}}$ to $\mathcal{V}$.
	\item  $\mathcal{V}$ checks that for $i\in \mathbf{ e}|_1$ that  \begin{enumerate}
		\item $\open(c_t^{(i)},t^{(i)},r_t^{(i)})=1$;
		\item $\open(c_L^{(i)},T_L^{(i)},r_L^{(i)})=1$; and 
		\item $T_L^{(i)}\neq f(t^{(i)})$.
	\end{enumerate} 
\end{enumerate}
Assume that $\mathcal{P}$ chose exactly one element of the backdoor in $\verificationkey$ wrongly, then this will be revealed by $\cnc$ to an honest $\mathcal{V}$ with probability $1/2$ (where $\mathcal{P}$ must open $\verificationkey|_1^{\mathbf{e}}$ to  the values he put into $c_t^{(i)}, c_L^{(i)}$ during $\keygen$ due to the binding-property of the commitment). In general, one can show that a cheating $\mathcal{P}$ can put at most $n$ non-backdooring inputs into $\verificationkey|_{0}^{\mathbf{ e}}$ except with probability negligible in $n$. Therefore, if the above check passes  for $\ell=4n$ at then least $1/2$ of the  values for $\verificationkey|_{0}^{\mathbf{ e}}$ must have the wrong  committed label as in a valid backdoor with overwhelming probability.

The above argument can  be made non-interactive and thus publicly verifiable using the Fiat-Shamir transform\cite{fiat_shamir}: in the protocol $\cnc$, $\mathcal{P}$ can generate the bit string $\mathbf{ e}$ itself by hashing $\verificationkey$ using a cryptographic hash function $H$. Then $\mathbf{ e}$ will be distributed as if it was chosen by an honest verifier, while it is sufficiently random by the guarantees of the hash function to allow the same analysis for cut-and-choose. Any $\mathcal{V}$ can recompute the value $\mathbf{ e}$ if it is generated from the commitments (while this also means that the challenge $\mathbf{ e}$ is generated after the commitments were computed), and we can turn the above algorithm $\cnc$ into the following non-interactive key-generation algorithm $\publickeygen$.
\paragraph*{$\publickeygen(\ell):$} ~
\begin{enumerate}
	\item Run $(\markingkey,\verificationkey)\leftarrow \keygen(\ell)$.
	\item Compute $\mathbf{e} \leftarrow H(\verificationkey)$.
	\item Set $\markingkey_p \leftarrow (\markingkey,\mathbf{ e})$, 
	 $\verificationkey_p \leftarrow (\verificationkey, \markingkey|_1^{\mathbf{ e}})$ and return $(\markingkey_p,\verificationkey_p)$.
\end{enumerate}

\subsubsection{Constructing the Public Verification Algorithm}
In the modified scheme, the  $\marking$ algorithm will only use the private subset $ \markingkey|_0^{\mathbf{ e}}$  of $\markingkey_p$ but will otherwise remain unchanged. The public verification algorithm for a model $M$ then follows the following structure:
\begin{enumerate*}[label=(\roman{*})]
	\item $\mathcal{V}$ recomputes the challenge $\mathbf{ e}$; 
	\item $\mathcal{V}$ checks $\verificationkey_p$ to assure that all of $\verificationkey|_1^{\mathbf{ e}}$ will form a valid backdoor ; and 
	\item $\mathcal{P},\mathcal{V}$ run $\classify$ on $\markingkey|_0^{\mathbf{ e}}$ using the interactive zero-knowledge argument system, and further test if the watermarking conditions on $M,\markingkey|_0^{\mathbf{ e}},\verificationkey|_0^{\mathbf{ e}}$ hold.
\end{enumerate*}

For an arbitrary model $M$, one can rewrite the steps \ref{step:verify:testcommitments} and \ref{step:verify:testclassify} of $\verify$ (using $M,\open,\classify$) into a binary circuit $C$ that outputs $1$ iff the prover inputs the correct $\markingkey|_0^{\mathbf{ e}}$ which opens  $\verificationkey|_0^{\mathbf{ e}}$ and if enough of these openings satisfy $\classify$. Both $\mathcal{P},\mathcal{V}$ can generate this circuit $C$ as its construction does not involve private information. For the interactive zero-knowledge argument, we let the relation $R$ be defined by boolean circuits that output $1$ where $x=C, w=\markingkey|_0^{\mathbf{ e}}$ in the following protocol $\publicverify$, which will obtain the model $M$ as well as $\markingkey_p = (\markingkey,\mathbf{ e})$ and  $\verificationkey_p =  (\verificationkey, \markingkey|_1^{\mathbf{ e}})$ where $\verificationkey= \{ c_t^{(i)}, c_L^{(i)} \}_{i \in [\ell]}$, $\markingkey = (\mathsf{b}, \{ r_t^{(i)}, r_L^{(i)} \}_{i \in [\ell]} )$ and $\mathsf{b}=\{ T^{(i)},T_L^{(i)}  \}_{i \in [\ell]}$ as input.
\begin{enumerate}
	\item\label{step:app:cnc}  $\mathcal{V}$ computes $\mathbf{ e}' \leftarrow H(\verificationkey)$. If $\markingkey|_1^{\mathbf{ e}}$ in $\verificationkey_p$ does not match $\mathbf{ e}'$ then abort, else continue assuming $\mathbf{ e}=\mathbf{ e}'$.
	\item $\mathcal{V}$ checks  that for all $i\in \mathbf{ e}|_1$:
		\begin{enumerate}
		\item $\open(c_t^{(i)},t^{(i)},r_t^{(i)})=1$ 
		\item $ \open(c_L^{(i)},T_L^{(i)},r_L^{(i)})=1$
		\item $ T_L^{(i)}\neq f(t^{(i)})$ 
	\end{enumerate} 
If one of the checks fails, then $\mathcal{V}$ aborts.
\item $\mathcal{P},\mathcal{V}$ compute a circuit $C$ with input $\markingkey|_0^{\mathbf{ e}}$	that outputs $1$ iff for all $i\in \mathbf{ e}|_0$:
\begin{enumerate}
\item $\open(c_t^{(i)},t^{(i)},r_t^{(i)})=1$ 
\item $\open(c_L^{(i)},T_L^{(i)},r_L^{(i)})=1$.
\end{enumerate}
Moreover, it tests that $\classify(t^{(i)},M)=T_L^{(i)}$ for all but $\epsilon|\mathbf{ e}|_0|$ elements.
	\item $\mathcal{P},\mathcal{V}$ run a zero-knowledge argument for the given relation R using $C$ as the statement, where the witness $\markingkey|_{0}^{\mathbf{e} }$ is the secret input of $\mathcal{P}$. $\mathcal{V}$ accepts iff the argument succeeds.
\end{enumerate}

Assume the protocol $\publicverify$ succeeds. Then in the interactive argument, $M$ classifies at least $(1-\epsilon)|\mathbf{ e}|_0|\approx(1-\epsilon)2n$ values of the backdoor $\mathsf{b}$ to the committed value. For $\approx n$ of the commitments, we can assume that the committed label does not coincide with the ground-truth function $f$ due to the guarantees of Step \ref{step:app:cnc}. It is easy to see that this translates into a $2\epsilon$-guarantee for the correct backdoor. By choosing a larger number $\ell$ for the size of the backdoor, one can achieve values that are arbitrarily close to $\epsilon$ in the above protocol.

\end{document}